\documentclass[11pt]{article}

\usepackage{microtype} % microtypography
\usepackage{booktabs}  % tables
\usepackage{url}  % urls

\usepackage{amsmath}
\usepackage{amsthm}

\usepackage[final,workshop]{automl}
\usepackage{automl}
\usepackage{tikz}
\usepackage{amsmath}
\usepackage{caption}
\usepackage{graphicx}

\newcommand{\mS}{\mathcal{S}}
\newcommand{\mA}{\mathcal{A}}
\newcommand{\mG}{\mathcal{G}}
\newcommand{\mM}{\mathcal{M}}

\newcommand{\na}{n_\text{act}}
\newcommand{\bR}{\mathbb{R}}
\usepackage{natbib}
\title{CANDID DAC: Leveraging Coupled Action Dimensions with Importance Differences in DAC}

\author[1]{\nameemail{Philipp Bordne}{bordnep@cs.uni-freiburg.de}}
\author[1]{\nameemail{M. Asif Hasan}{hasan@cs.uni-freiburg.de}}
\author[1]{\nameemail{Eddie Bergman}{bergmane@cs.uni-freiburg.de}}
\author[1]{\nameemail{Noor Awad}{awad@cs.uni-freiburg.de}}
\author[1]{\nameemail{André Biedenkapp}{biedenka@cs.uni-freiburg.de}}

\affil[1]{University of Freiburg}

\hypersetup{%
  pdfauthor={Philipp Bordne}, % will be reset to "Anonymous" unless the "final" package option is given
  pdftitle={CANDID DAC: Leveraging Coupled Action Dimensions with Importance Differences in DAC},
  pdfsubject={Dynamic Algorithm Configuration},
  pdfkeywords={Dynamic Algorithm Configuration, Reinforcement Learning, high-dimensional action spaces}
}

\begin{document}

\maketitle

\begin{abstract}
    High-dimensional action spaces remain a challenge for dynamic algorithm configuration (DAC). 
    Interdependencies and varying importance between action dimensions are further known key characteristics of DAC problems. We argue that these Coupled Action Dimensions with Importance Differences (CANDID) represent aspects of the DAC problem that are not yet fully explored. To address this gap, we introduce a new white-box benchmark within the DACBench suite that simulates the properties of CANDID. Further, we propose sequential policies as an effective strategy for managing these properties. Such policies factorize the action space and mitigate exponential growth by learning a policy per action dimension. At the same time, these policies accommodate the interdependence of action dimensions by fostering implicit coordination. We show this in an experimental study of value-based policies on our new benchmark. This study demonstrates that sequential policies significantly outperform independent learning of factorized policies in CANDID action spaces. In addition, they overcome the scalability limitations associated with learning a single policy across all action dimensions. The code used for our experiments is available under \url{https://github.com/PhilippBordne/candidDAC}.
\end{abstract}

\section{Introduction}
In the Dynamic Algorithm Configuration (DAC) problem \citep{Biedenkapp2020DynamicAC} hyperparameters must be adjusted on-the-fly.
A significant portion of these are either categorical or discrete, posing a challenge for reinforcement learning (RL), the common approach to solving DAC \citep{adriaensen2022automated}, as they result in a combinatorial explosion of the joint action space. The DAC problem is further complicated by interaction effects between hyperparameters \citep{2014_hutter_fanova, van_Rijn_2018, 2023_interact_lr_bs} with varying importance of hyperparameters \citep{2014_hutter_fanova, moosbauer2021pdp_hpo, mohan2023autorl, 2018cave}. In this paper, we will refer to these properties as Coupled ActioN-Dimensions with Importance Differences (CANDID). This work investigates how CANDID influences the performance of RL algorithms and lays the ground for the development of better methods to tackle the DAC problem. 
To do so, we introduce a new CANDID benchmark derived from the original Sigmoid benchmark \citep{Biedenkapp2020DynamicAC}. We believe that this captures the complexities associated with the high-dimensionality of action spaces in DAC and other control domains more comprehensively.
We use this benchmark to evaluate RL algorithms that learn policies per action dimension in factored action spaces \citep{sharma2017learning, xue2022multiagent}.
In particular, we implement two algorithm variants to learn sequential policies \citep{metz2019discrete}. We compare them against a single agent baseline as typically used in DAC as well as a multi-agent baseline. Our results suggest that under the CANDID properties, sequential policies can coordinate action selection between dimensions while avoiding combinatorial explosion of the action space. Our initial results encourage an extended study of sequential policies for DAC.

\section{Related Work}
\paragraph{Hyperparameter importances and interactions in the AutoML landscape} 
Hyperparameter importance is a key topic in AutoML. Tools like fANOVA \citep{2014_hutter_fanova} quantify the importance of individual hyperparameters and their interactions. These tools guide the tuning of hyperparameters and the examination of pipeline components in post-hoc analyses \citep{van_Rijn_2018}. To our knowledge, the importance and interactions of hyperparameters have not yet been used as structural information to solve DAC problems more effectively.
\paragraph{Solving high-dimensional action spaces in RL} Large discrete or categorical action spaces pose a significant challenge for RL, as well as DAC by RL \citep{biedenkapp2022gecco}. Factored action space representations (FAR) address this challenge by learning policies per action dimension \citep{sharma2017learning}. Based on Deep Q-Networks (DQN) \citep{mnih2015dqn}, \citet{metz2019discrete} introduced sequential policies that control one action dimension at a time and condition on previously selected actions. This approach has been extended to more RL-algorithms \citep{pierrot2021factored}. We further develop this concept by ordering the sequential policies based on the importance of action dimensions and propose an adaptation of its training algorithm inspired by sequential games.
Other research has approached high-dimensional action spaces as multi-agent learning problems, where each action dimension is controlled by a distinct agent.
For example, \citet{xue2022multiagent} trained individual agents per hyperparameter to dynamically tune a multi-objective optimization algorithm.
In their study, well-established MARL algorithms such as VDN \citep{vdn} and QMIX \citep{qmix} were utilized to coordinate learning in the multi-agent setting. As we aim to explicitly exploit the CANDID property of DAC action spaces through sequential action selection, we focus solely on the simplest MARL baseline which is independent learning of agents. Analogous to MARL research, allowing sequential policies to observe previously selected actions acts as a communication scheme between agents. This approach is tailored to the structure revealed by the importance of action dimensions and their interaction effects. For further insights into the state-of-the-art of MARL, we refer interested readers to the survey of \citet{huh2023multiagent}.

\section{Piecewise Linear Benchmark} \label{sec:benchmark}
\begin{figure}[h]
  \centering
  \begin{subfigure}[b]{0.49\textwidth}
  \centering
    \includegraphics[width=\textwidth]{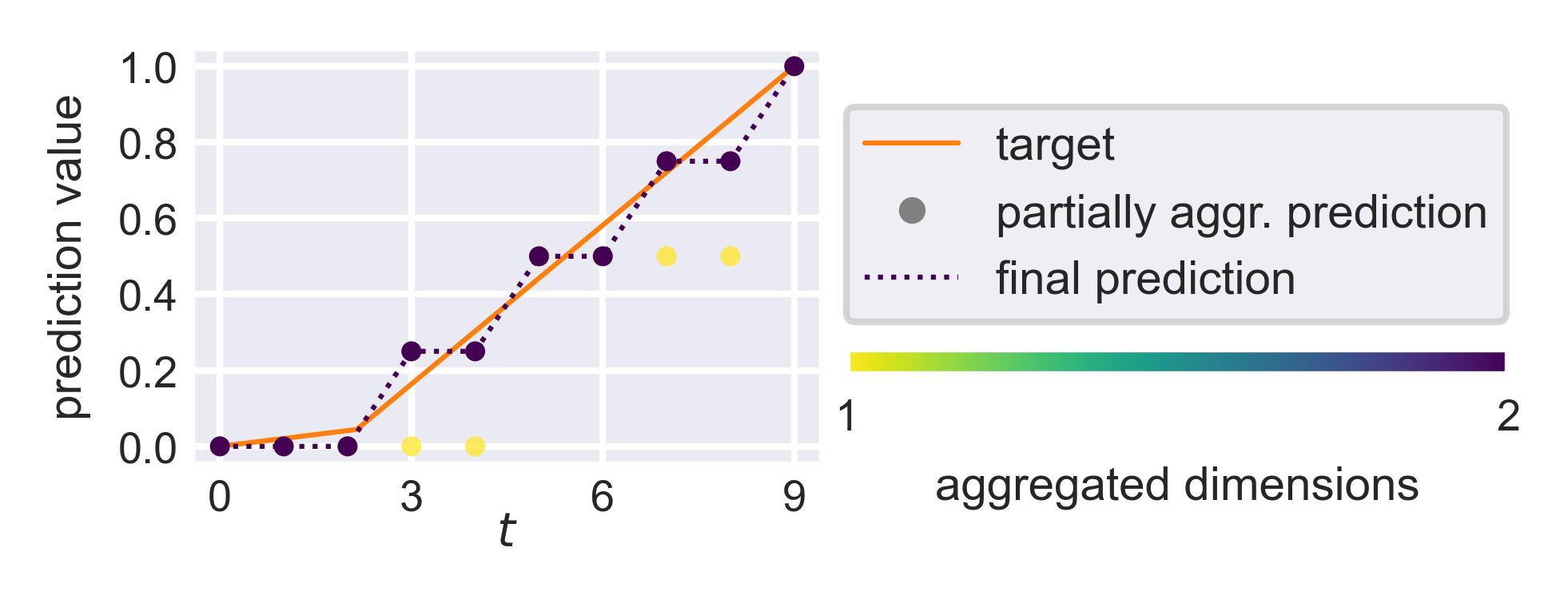}
  \caption{} \label{fig:bench_ex}
  \end{subfigure}\hfill
  \begin{subfigure}[b]{0.49\textwidth}
  \centering
    \includegraphics[width=0.9\textwidth]{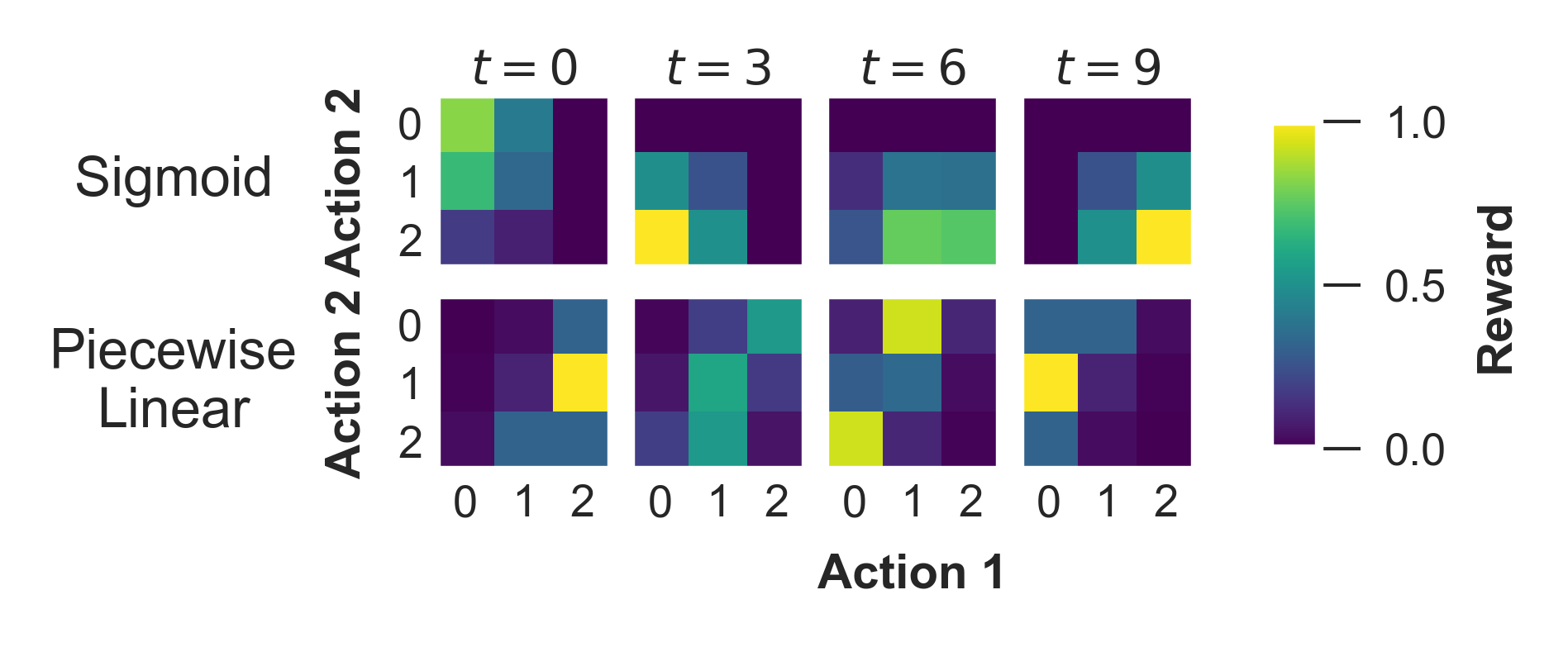}
  \caption{}\label{fig:rew_surf}
  \end{subfigure}
\caption{(a) Example of a prediction task on a 2D Piecewise Linear instance; (b) Comparison of its reward surface against a 2D Sigmoid instance at 4 different time steps.}
\end{figure}

\noindent We implement the Piecewise Linear benchmark within DACBench \citep{eimer2021dacbench}, building on the Sigmoid benchmark that models a DAC problem with a high-dimensional action space \citep{Biedenkapp2020DynamicAC}.
In a $M$-D Sigmoid, the task is to predict the values of $M$ individual Sigmoid curves over $T=10$ time steps, with $\na^m$ action choices for dimension $m$. Its reward function is defined as $r_t = \prod_{m=1}^M 1 - \texttt{pred\_error}(a^m_t)$, which can be maximized by minimizing the prediction errors independently per action dimension. In contrast the Piecewise Linear benchmark aggregates all action dimensions through a weighted sum to predict on a \textit{single} target function (Figure \ref{fig:bench_ex}):
\begin{align}
    \texttt{pred}(a_t^{1:M}) = \frac{a^1_t}{\na^1 - 1} + \sum_{m=2}^M w_m(\frac{a^m_t}{\na^m - 1} - \frac{1}{2}) \label{eq:aggreg}
\end{align}
This introduces coupling and importance differences between action dimensions, emulating the CANDID setting, and requires joint optimization of action dimensions (Figure \ref{fig:rew_surf}).
Instead of Sigmoid curves, we sample 2-segment piecewise linear functions as prediction targets to avoid constant target function values featured by many Sigmoid instances. We present more details such as reward signal definition and the train and test dataset in Appendix \ref{app:pl}.

\section{Controlling CANDID Action Spaces with Sequential Policies}
The aim of RL is to learn the optimal policy for a Markov Decision Process (MDP) $\mathcal{M} = (\mathcal{S}, \mathcal{A}, P, R)$ \citep{SuttonBarto18}. After factorizing a $M$-dimensional action space $\mA = \mA_1 \times ... \times \mA_M$ into $1$-dimensional action spaces $\mA_1, ... , \mA_M$ we can learn a policy per action dimension. We will refer to such approaches as factorized policies. Sequential policies build on the idea of extending the original MDP $\mM$ to a sequential MDP (sMDP) by introducing substates $([s_t,],[s_t,a_t^1],...,[s_t,a_t^{1:M-1}])$ to include an action selection \textit{process} for all action dimensions at the current time step $t$ \citep{metz2019discrete}. A comprehensive formalization of these MDP reformulations is given in Appendix \ref{app:detail_mdp}.

We choose sequential policies to solve CANDID action spaces for two reasons. (I) Sequential policies are able to condition subsequent actions on already selected actions and thus to learn about the coupling between action dimensions. (II) The different importances of action dimensions induce an order for the selection process: By selecting the most important action first, this information is available when controlling all other action dimensions at the current time step.
We implemented two different algorithms to learn sequential policies, which differ in the TD-updates (see, Equation \eqref{eq:td_saql_sdqn}). The first is a simplified version of SDQN (simSDQN) that omits the upper Q-network compared to the implementation of \cite{metz2019discrete}. This approach can be interpreted as a hybrid of tabular and function-approximation Q-learning. We introduce an implicit state that denotes the current stage in the action selection process and add a tabular entry for each of the $M$ stages. The second approach can be interpreted as multiple agents playing a sequential game (Sequential Agent Q-Learning = SAQL) and selecting $a^m_t$ is the turn of the $m$-th agent, an approach inspired by learning equilibria in Stackelberg games \citep{gerstgrasser2023oracles}. As our setting is fully cooperative, all agents receive the same true reward. We provide a more detailed explanation in Appendix \ref{app:dqn_in_mdp}.
\begin{subequations}\label{eq:td_saql_sdqn}
    \begin{align}
        \text{target}^m_{\text{SAQL}} & = r_t + \gamma \max_{a^m} Q^m([s_{t+1}, a_{t+1}^{1:m-1}], a^m) = r_t + \gamma V^m(s^m_{t+1}) \\
        \text{target}^m_{\text{simSDQN}} & = \begin{cases}
                                        \max_{a^{m+1}} Q^{m+1}([s_t, a^{1:m}_t], a^{m+1}) = V^{m+1}(s^{m+1}_t), & \quad 1 \leq m \leq M-1 \\
                                        r_t + \gamma \max_{a^1} Q^{1}([s_{t+1},], a^1) = r_t + \gamma V^{1}(s^{0}_{t+1}), & \quad m = M
                                  \end{cases}
    \end{align}
\end{subequations}

\section{Experimental Setup} \label{sec:setup}
We compared sequential policies on the Sigmoid and Piecewise Linear benchmark against two baselines: (I) Double DQN (DDQN) \citep{vanhasselt2015ddqn} as baseline of a single-agent policy controlling all action dimensions simultaneously; (II) Independent Q-Learning (IQL) \citep{tampuu2015iql} as multi-agent baseline controlling all action dimensions independently. Note that IQL can be seen as an ablation of SAQL without communication between individual agents. We provide a detailed comparison of the evaluated algorithms in Appendix \ref{app:dqn_in_mdp}.

Additional action dimensions of the Piecewise Linear benchmark enable higher rewards through more accurate predictions. As static baseline, we calculated the reward achievable by predictions based solely on the first and most important action dimension (\textit{optimal (1D)}). Outperforming this baseline indicates effective coordination of action dimensions.

We selected hyperparameters on a per-algorithm basis on the training instance set of the $5$D Sigmoid benchmark. To this end, we evaluated each algorithm on a portfolio of $100$ randomly sampled hyperparameter configurations. Each evaluation consisted of $10$ random seeds that we aggregated through the median \citep{agarwal2021rl_statistics}. We provide the configurations together with the model architectures in Appendices \ref{app:hyperparams} \& \ref{app:policy_architecture} and the used computing resources in Appendix \ref{app:resources}.

To induce importance differences, we define the weights per action dimension as  $w_m = \lambda^{m-1}$ for $m\in\{2, ... , M\}$. We set the importance decay $\lambda = 0.5$ for all experiments and report the results for further importance decays in Appendix \ref{app:diff_importance}.
To assess scaling with action space size, we evaluated the approaches on varying action space dimensionality (\texttt{dim}, or $M$) and actions per dimension (\texttt{n\_act}).
We provide details on the train and test dataset in Appendix \ref{app:pl}.

\section{Results and Discussion} \label{sec:results}

\begin{figure}[h]
    \centering
    \includegraphics[]{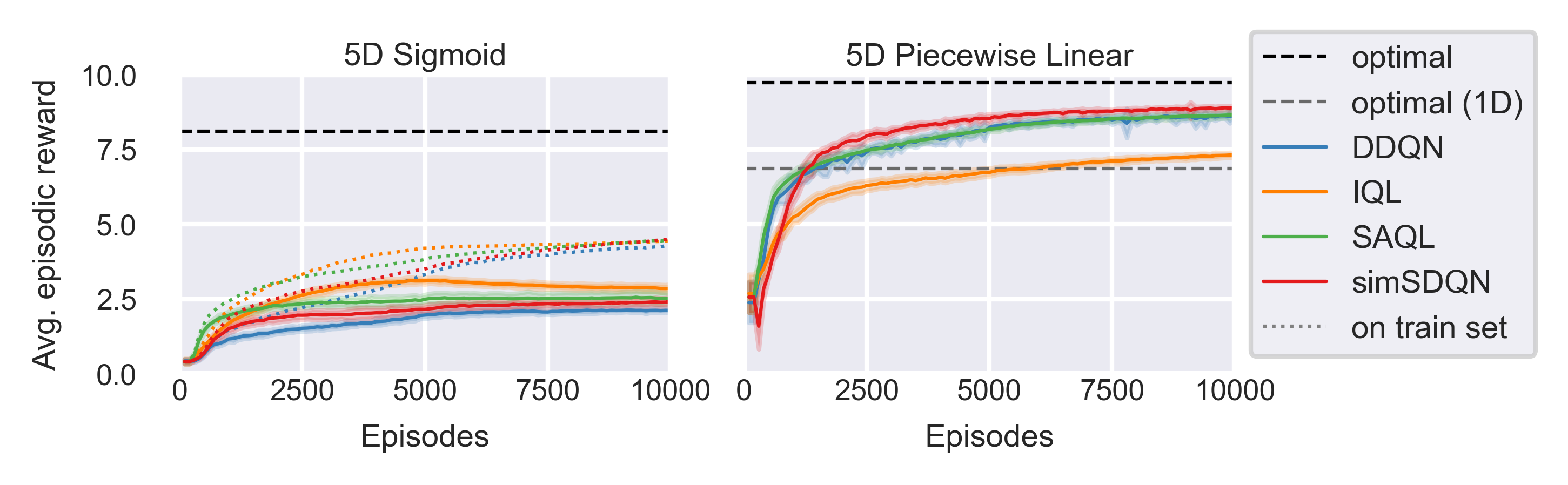}
    \caption{Average episodic rewards (mean, std from 20 seeds) on the test sets of 5D Sigmoid and Piecewise Linear benchmark ($\lambda = 0.5$ and $\texttt{n\_act} = 3$). Generalization error on 5D Sigmoid results from hyperparameter selection on its training set.}
    \label{fig:sig-vs-pl}
\end{figure}

\textbf{Are Sequential Policies Beneficial for CANDID DAC?\phantom{.}}
To answer this question, we compare the best performing approaches on the 5D Sigmoid without CANDID properties and the 5D Piecewise Linear benchmark with CANDID properties.
The performance of the multi-agent baselines (IQL) is notably poor in the CANDID setting and performs best in other scenarios, as illustrated in Figure \ref{fig:sig-vs-pl}. Even on the 2D Piecewise Linear benchmark, where all other evaluated algorithms achieve near-perfect solutions within a fraction of the total training episodes, IQL lags behind. This demonstrates the need for a mechanism of coordination between action dimensions with interaction effects and becomes especially evident when comparing performances of IQL and SAQL.

\begin{figure}
    \centering
    \includegraphics[width=\textwidth]{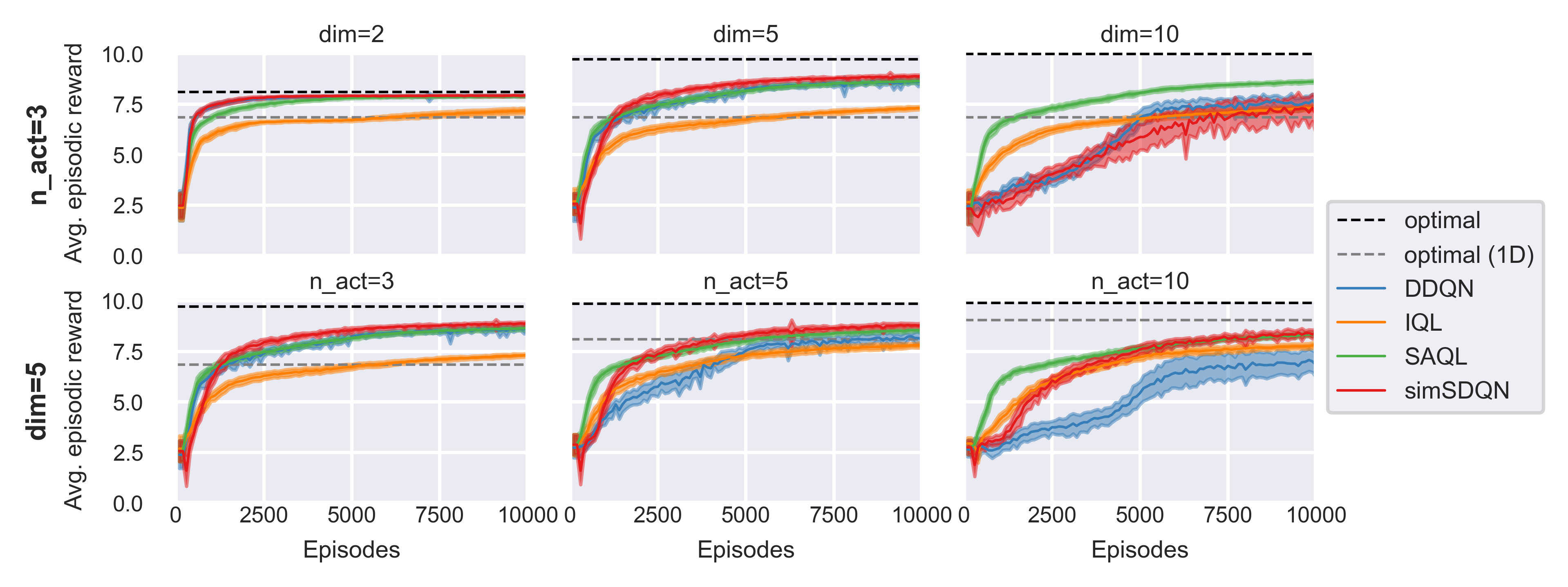}
    \caption{Experiments investigating scaling behavior of algorithms. First row keeps number of actions per action dimension fixed at $\texttt{n\_act}=3$ and varies dimensionality $\texttt{dim}$ of action space. Second row keeps $\texttt{dim}$ fixed and varies $\texttt{n\_act}$. Both experiments keep importance decay $\lambda=0.5$ fixed. Rewards from test set (mean, std from 20 seeds).}
    \label{fig:dim_n_act_scaling}
\end{figure}

\noindent\textbf{Can Sequential Policies Scale to Larger Configuration Spaces?\phantom{.}}
To answer this, we vary the number of discrete action choices per dimension (n\_act) and the number of action dimensions.
Although being slightly outperformed by simSDQN and matched by DDQN on the 5D Piecewise Linear benchmark, the performance of SAQL (as IQL) remains stable when increasing the dimension of the action space to $10$ (Figure \ref{fig:dim_n_act_scaling}, upper row). simSDQN is notably impacted by the expansion of action space dimensionality. This outcome is likely attributed to the reward signal being solely observed in the TD-update of the last, i.e. the least important, action dimension. As a result, reward information must be propagated through more Q-networks before it reaches the most important policies, which is required for them to identify their optimal actions. DDQNs performance also decreases, as adding action dimensions leads to exponential growth of its action space. Increasing the number of discrete action choices per action dimension has a similar effect on DDQN as it still results in polynomial growth in size of its action space. Factorized policies are not negatively affected by this change, as can be seen in the bottom row of Figure \ref{fig:dim_n_act_scaling}, as it results only in a linear growth in size of their action spaces.

We further observed that the degree of importance differences between action dimensions has minimal impact on the relative performances of the evaluated algorithms, as detailed in Appendix \ref{app:diff_importance}. In another ablation, we found that inverting the importance order of action selection in sequential policies slightly reduces their performance, which supports our initial intuition about the effectiveness of selecting the most important action first (Appendix \ref{app:reverse_importances}).

\section{Limitations} \label{sec:limitations}
This exploratory study is limited by its reliance on a white-box benchmark emulating the CANDID properties. It may not fully encapsulate the complexity of real-world applications, which could limit the general applicability of the findings. It is also limited to the most basic multi-agent baseline, IQL and comparison against more advanced algorithms such as QMIX would allow a better contextualization of sequential policies' performance within the SOTA.
Several of the evaluated algorithms achieved near optimal performance. This might limit its suitability for the validation of improvements to these algorithms. A major drawback of our implementation of sequential policies as compared to traditional multi-agent algorithms is that the dimensionality of their observation space grows linearly with the number of interacting action dimensions and it is not straightforward to add policies later on.

\section{Broader Impact Statement} \label{sec:impact}
While this work was specifically directed towards DAC, the challenge of controlling numerous interacting inputs is ubiquitous in most complex real-world systems. As such, the introduced approach could be applied in settings where communication between components of these systems is possible. Some examples are robots with multiple joints, smart buildings or grids. Application of our approach would require prior analysis of the structure of the control problem, in particular, to identify interacting control inputs and their relative importance. Sequential policies require, like other factorized approaches, the learning of multiple policies. This can lead to an increased demand for training and computing time and memory.

\section{Conclusions}
In this report, we introduced coupled action dimensions with importance differences (CANDID) as an under-researched challenge for reinforcement learning \& DAC and provided the Piecewise Linear benchmark for assessing RL algorithms in these settings. In our experimental study, we have shown that sequential policies are a promising technique to address this challenge. In future work, we plan to apply SAQL to real-world inspired benchmarks and to compare it against more advanced MARL baselines that use value function factorization such as QMIX. As we see value function factorization to be mainly orthogonal to SAQL, we also plan to combine the two approaches. Moreover, we plan to improve agent coordination by exploring learned message passing, inspired by existing approaches like \cite{huang2020policy}. This strategy aims to prevent observation space growth with action dimensionality, improving scalability.

% The 9 pages allocated for the main paper must include a broader impact
% statement regarding the approach, datasets and applications proposed/used in
% your paper. It should reflect on the environmental, ethical and societal
% implications of your work. The statement should require at most one page and
% must be included both at submission and camera-ready time.
%
% If authors have reflected on their work and determined that there are no
% likely negative broader impacts, they may use the following statement:
%
% After careful reflection, the authors have determined that this work presents
% no notable negative impacts to society or the environment.
%
% This section is included in the template as a default, but you can also place these
% discussions anywhere else in the main paper, e.g., in the introduction/future work.
%
% The Centre for the Governance of AI has written an excellent guide for writing
% good broader impact statements (for the NeurIPS conference) that may be a
% useful resource for AutoML-Conf authors:
%
% https://medium.com/@GovAI/a-guide-to-writing-the-neurips-impact-statement-4293b723f832
\newpage
\begin{acknowledgements}
The authors acknowledge support by the state of Baden-Württemberg through bwHPC. Noor Awad and André Biedenkapp acknowledge
funding from The Carl Zeiss Foundation through the research network “Responsive and Scalable
Learning for Robots Assisting Humans” (ReScaLe) of the University of Freiburg.
We acknowledge funding by the Deutsche Forschungsgemeinschaft (DFG, German Research Foundation) under SFB 1597 (SmallData), grant number 499552394.
\end{acknowledgements}

%%%%%%%%%%%%%%%%%%%%%%%%%%%%%%%%%%%%%%%%%%%%%%%%%%%%%%%%
\bibliography{literature}
%%%%%%%%%%%%%%%%%%%%%%%%%%%%%%%%%%%%%%%%%%%%%%%%%%%%%%%%

\section*{Submission Checklist}

\begin{enumerate}
\item For all authors\dots
  \begin{enumerate}
  \item Do the main claims made in the abstract and introduction accurately
    reflect the paper's contributions and scope?
    \answerYes{The benchmark design reflects the stated challenges (section \ref{sec:benchmark}) and experimental results and discussion show superior performance of our approach (section \ref{sec:results})}
  \item Did you describe the limitations of your work?
    \answerYes{see section \ref{sec:limitations}}
  \item Did you discuss any potential negative societal impacts of your work?
    \answerYes{see section \ref{sec:impact}}
  \item Did you read the ethics review guidelines and ensure that your paper
    conforms to them? \url{https://2022.automl.cc/ethics-accessibility/}
    \answerYes{} We did several proofreads and use colorblind-friendly colors in our plots.
  \end{enumerate}
\item If you ran experiments\dots
  \begin{enumerate}
  \item Did you use the same evaluation protocol for all methods being compared (e.g.,
    same benchmarks, data (sub)sets, available resources)?
    \answerNo{} We used the same benchmarks and instance sets for all methods. We trained all methods on the same type of CPU but for the DDQN experiments with (dim=10, n\_act=3) and (dim=5, n\_act=10) we had to use a GPU due to the size of the resulting Q-networks. However this shouldn't affect our comparisons since we compared progress over the number of episodes and also limited training by the number of episodes and not in training time.
  \item Did you specify all the necessary details of your evaluation (e.g., data splits,
    pre-processing, search spaces, hyperparameter tuning)?
    \answerYes{} We describe the hyperparameter search process qualitatively in section \ref{sec:setup} and the used hyperparameters in appendix \ref{app:hyperparams}. The benchmark and its instances are part of the published code, the search for hyperparameters is conducted through the main experimental script and comprehensible from there.
  \item Did you repeat your experiments (e.g., across multiple random seeds or splits) to account for the impact of randomness in your methods or data?
    \answerYes{} We evaluated every (method,benchmark)-pair on 20 random seeds. During hyperparameter selection we evaluated every method on 10 random seeds per configuration.
  \item Did you report the uncertainty of your results (e.g., the variance across random seeds or splits)?
    \answerYes{} We included std in our plots.
  \item Did you report the statistical significance of your results?
    \answerNo{} This study is exploratory and we believe our conclusions are justifiable by the plots including std.
  \item Did you use tabular or surrogate benchmarks for in-depth evaluations?
    \answerNo{} This was not in our scope of finding suitable RL algorithms for DAC.
  \item Did you compare performance over time and describe how you selected the maximum duration?
    \answerNo{} We did no such comparisons.
  \item Did you include the total amount of compute and the type of resources
    used (e.g., type of \textsc{gpu}s, internal cluster, or cloud provider)?
    \answerYes{} see appendix \ref{app:resources}
  \item Did you run ablation studies to assess the impact of different
    components of your approach?
    \answerYes{} We did limited ablations: comparing IQL and SAQL (section \ref{sec:results}) and by validating the order of sequential policies (appendix \ref{fig:reverse_imp})
  \end{enumerate}
\item With respect to the code used to obtain your results\dots
  \begin{enumerate}
\item Did you include the code, data, and instructions needed to reproduce the
    main experimental results, including all requirements (e.g.,
    \texttt{requirements.txt} with explicit versions), random seeds, an instructive
    \texttt{README} with installation, and execution commands (either in the
    supplemental material or as a \textsc{url})?
    \answerYes{} We provided a link to the repository.
  \item Did you include a minimal example to replicate results on a small subset
    of the experiments or on toy data?
    \answerYes{} Most of the experiments are small-scale and can be executed locally. We provide examples how to execute them locally.
  \item Did you ensure sufficient code quality and documentation so that someone else
    can execute and understand your code?
    \answerYes{} We use hydra to define experimental setups and provide a description on how to that in the README. We also focused on documentation of the main script. The source code for our benchmark and algorithms is mostly typed but we are still working on improving it until publication.
  \item Did you include the raw results of running your experiments with the given
    code, data, and instructions?
    \answerYes{} The metrics in the plots are included in a .csv in the directory analysis/run\_data within the repository.
  \item Did you include the code, additional data, and instructions needed to generate
    the figures and tables in your paper based on the raw results?
    \answerYes{} In the directory analysis/ we provide the notebooks used to generate the plots.
  \end{enumerate}
\item If you used existing assets (e.g., code, data, models)\dots
  \begin{enumerate}
  \item Did you cite the creators of used assets?
    \answerYes{} We used DACBench and cite it in our paper
  \item Did you discuss whether and how consent was obtained from people whose
    data you're using/curating if the license requires it?
    \answerNA{} No such material was used.
  \item Did you discuss whether the data you are using/curating contains
    personally identifiable information or offensive content?
    \answerNA{} No such data was used.
  \end{enumerate}
\item If you created/released new assets (e.g., code, data, models)\dots
  \begin{enumerate}
    \item Did you mention the license of the new assets (e.g., as part of your code submission)?
    \answerYes{} license in repository
    \item Did you include the new assets either in the supplemental material or as
    a \textsc{url} (to, e.g., GitHub or Hugging Face)?
    \answerYes{} availabe on GitHub
  \end{enumerate}
\item If you used crowdsourcing or conducted research with human subjects\dots
  \begin{enumerate}
  \item Did you include the full text of instructions given to participants and
    screenshots, if applicable?
    \answerNA{}
  \item Did you describe any potential participant risks, with links to
    Institutional Review Board (\textsc{irb}) approvals, if applicable?
    \answerNA{}
  \item Did you include the estimated hourly wage paid to participants and the
    total amount spent on participant compensation?
    \answerNA{}
  \end{enumerate}
\item If you included theoretical results\dots
  \begin{enumerate}
  \item Did you state the full set of assumptions of all theoretical results?
    \answerNA{}
  \item Did you include complete proofs of all theoretical results?
    \answerNA{}
  \end{enumerate}
\end{enumerate}

\newpage
%%%%%%%%%%%%%%%%%%%%%%%%%%%%%%%%%%%%%%%%%%%%%%%%%%%%%%%%%%%%%%%%%%%%%%
\appendix

\section{Details on Piecewise Linear Benchmark} \label{app:pl}
\begin{figure} [h]
    \centering
    \includegraphics[width=\textwidth]{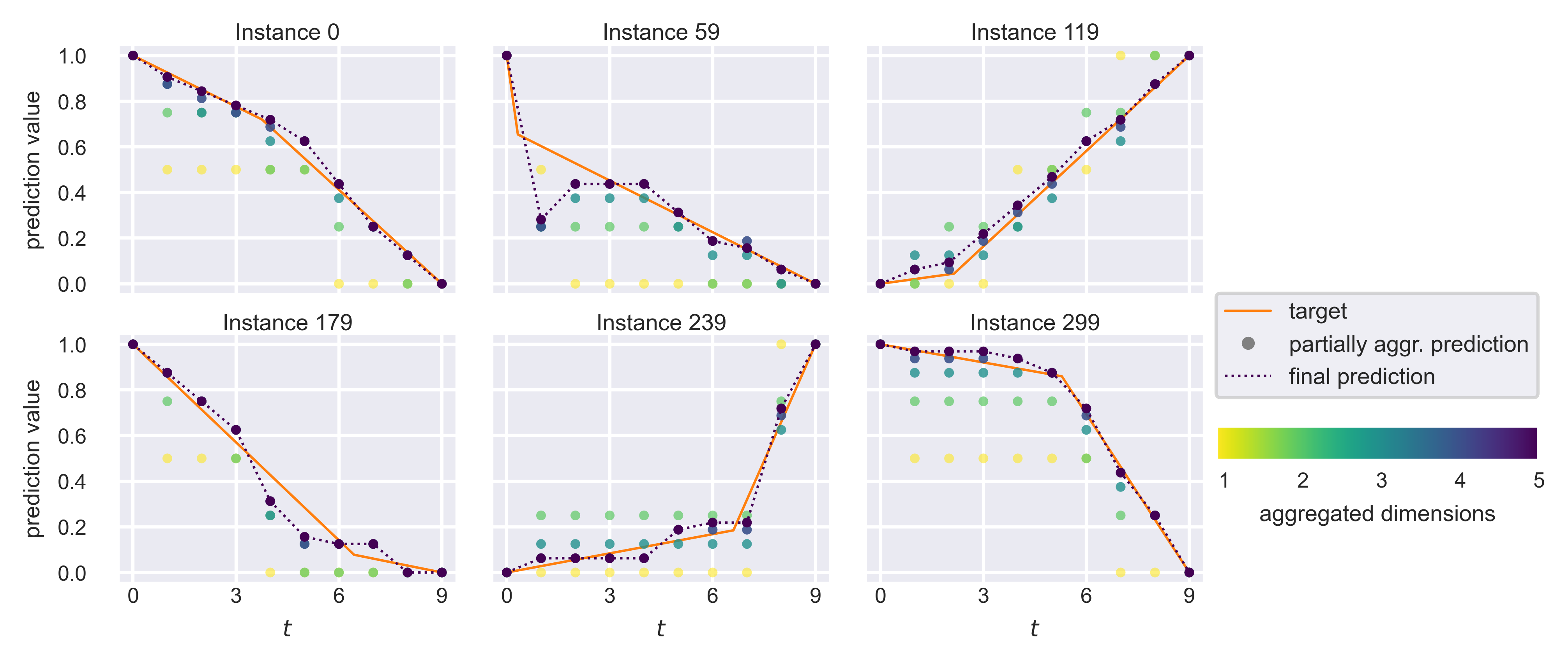}
    \caption{Several examples of test instances of the Piecewise Linear benchmark and predictions from a policy learned through SAQL.}
    \label{fig:pl_saql}
\end{figure}
\begin{figure} [h]
    \centering
    \includegraphics[width=\textwidth]{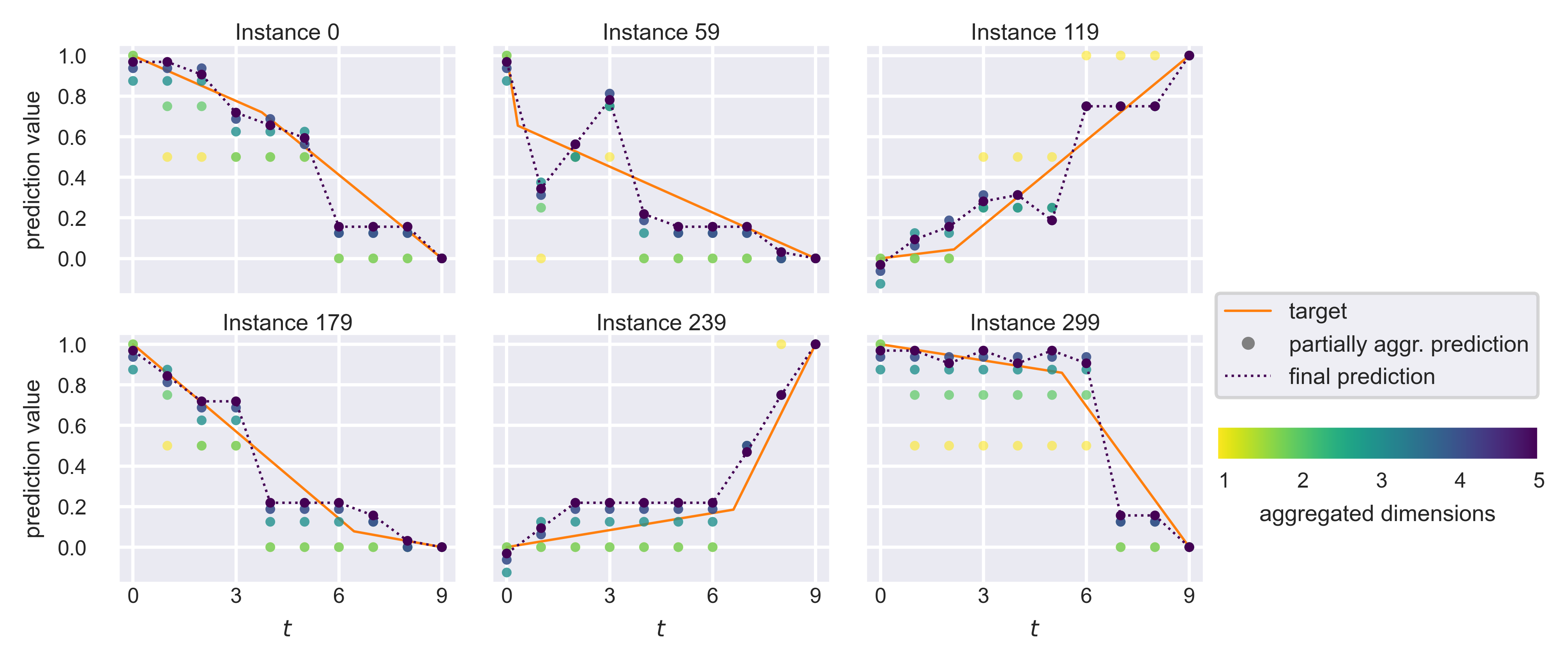}
    \caption{Several examples of test instances of the Piecewise Linear benchmark and predictions from a policy learned through IQL.}
    \label{fig:pl_iql}
\end{figure}
\paragraph{Train and Test Instance Datasets} The Piecewise Linear benchmark's training and test datasets $\mathcal{I}_\text{train}$ and $\mathcal{I}_\text{test}$ consist of 300 different target function instances each. Instances for both datasets were generated in a two-step process. First we randomly sample an intermediate point $(x, y)$, where $x \in [0, 9]$ and $y \in [0, 1]$. Next, a random choice determines whether to establish a connection between $(0, 0)$ and $(9, 1)$ (increasing) or between $(0, 1)$ and $(9, 0)$ (decreasing) through the sampled intermediate point. Together this defines the 2 piecewise linear segments of our target function. When interacting with the Piecewise Linear benchmark, one episode consists of predicting the function values of a specific target function instance over $T=10$ time steps. During training we reset the environment after finishing an episode and continue with predicting on the next target function instance (round-robin). We note that this defines a contextual MDP (cMDP) as it is proposed for the DAC framework \citep{Biedenkapp2020DynamicAC}. We also note that the described instances are independent from the number of action dimensions and actions per action dimension. Hence we used the same train and test instance datasets for all our experiments.

\begin{figure} [h]
    \centering
    \includegraphics[width=0.5\textwidth]{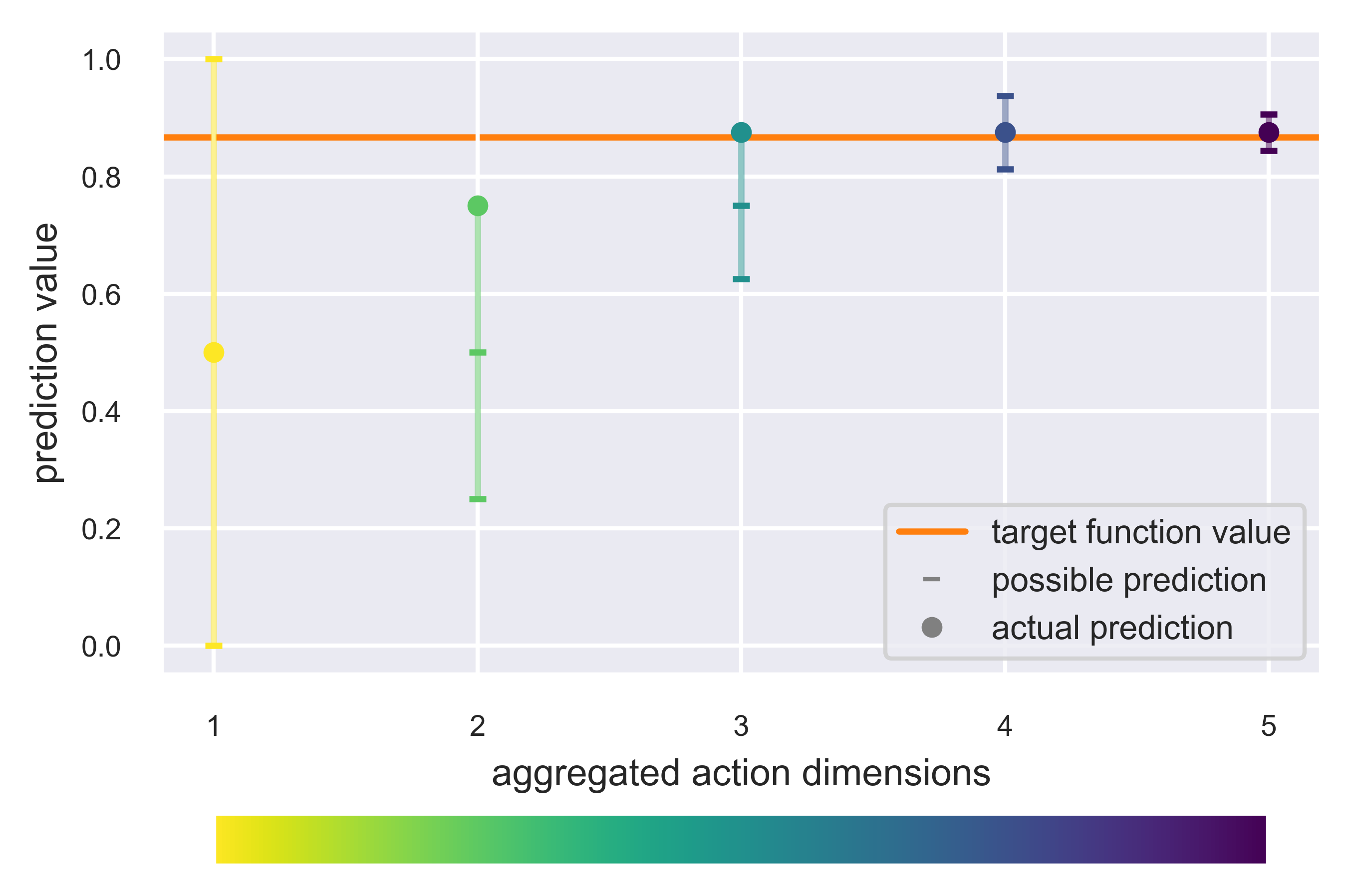}
    \caption{Illustration of action selection by a sequential policy trained through SAQL on the Piecewise Linear benchmark at time step $t=5$ of benchmark instance 299. It demonstrates how the prediction gets iteratively fine-tuned to fit the target value.}
    \label{fig:single_aggregate}
\end{figure}

Figure \ref{fig:pl_saql} visualizes different instances from the generated instance set, along with the resulting predictions obtained by aggregating an increasing number of action dimensions through a sequential policy. It shows that as more action dimensions are aggregated, the prediction accuracy generally improves. Figure \ref{fig:single_aggregate} illustrates, for a single time step, how a sequential policy progressively approximates the value of the prediction target more accurately as it aggregates more dimensions. In contrast, Figure \ref{fig:pl_iql} shows how independent policies often fail to coordinate.
\paragraph{Observation Space} Similar to the Sigmoid benchmark \citep{Biedenkapp2020DynamicAC} the observation space consists of information about the current instance $i \in \mathcal{I}$, the remaining number of steps for the current instance $T-t$ and the actions selected at the previous time step. The current instance is uniquely defined by the coordinates $(x,y)$ of the intermediate point and a bit $b$ which determines whether the target function is increasing or decreasing. Thus, the observation vector of policies in a single policy setting (DDQN) and in a factorized but non-sequential policy setting (IQL) is defined as $o_t = [T-t, x_i, y_i, b_i, a^1_{t-1}, ... , a^M_{t-1}]$. The observation of an atomic policy $m$ in the sequential policy setting (SAQL, simSDQN) additionally includes actions selected by its predecessors at the current time step: $o^m_t = [T-t, x_i, y_i, b_i, a^1_{t-1}, ... , a^M_{t-1}, a^1_{t}, ... , a^{m-1}_{t}]$.

\paragraph{Reward Computation} To compute the reward at step $t$, we first calculate the prediction error as the absolute difference between the aggregated prediction $\texttt{pred}(a_t^{1:M})$ and the value of the piecewise linear function $\texttt{pl}(t)$ at the current point in time:
\begin{align}
    \texttt{pred\_error}(a_t^{1:M}) = |\texttt{pred}(a_t^{1:M}) - \texttt{pl}(t)|
\end{align}
The aggregated prediction $\texttt{pred}(a_t^{1:M})$ is defined in Equation \eqref{eq:aggreg}. To incentivize learning across all action dimensions, including those with less significant contributions, we define an exponentially decaying reward signal $r_t = e^{-c \cdot \texttt{pred\_error}(a_t^{1:M})}$. This formulation ensures that selecting the first action optimally is necessary to obtain high rewards, but it might not be sufficient, as the exponentially decaying reward puts more emphasis on a \textit{precise} prediction. For our experiments, we have chosen $c=4.6$.

\newpage

\section{MDP Reformulations and Algorithms in Detail} \label{app:detail_mdp_algo}
%%%%%%%%%%%%%%%%%%%%%%%%%%%%%%%%%%%%%%%%%%%%%%%%%%%%%%%%%%%%%%%%%%%%%%%%%%%%%%%%%%%%%%%%
% MDP graphs
%%%%%%%%%%%%%%%%%%%%%%%%%%%%%%%%%%%%%%%%%%%%%%%%%%%%%%%%%%%%%%%%%%%%%%%%%%%%%%%%%%%%%%%%
\subsection{Action Space Factorization and MDP Reformulations} \label{app:detail_mdp}
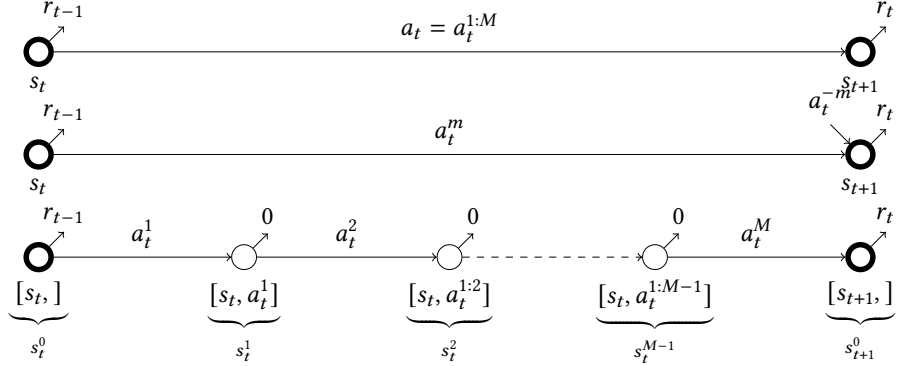
\begin{figure} [!h]
\centering
\resizebox{0.8\textwidth}{!}{
    \begin{tikzpicture}
        %%%%%%%%%%%%%%%%%%%%%%%%%%%%%%%%%%%
        % the original MDP
        \node[circle,draw, line width=0.8mm] (supnode1) at (0,3) {};
        \node[circle,draw, line width=0.8mm] (supnode5) at (12,3) {};
    
        % labels
        \node[below] at (supnode1.south) {$s_t$};
        \node[below] at (supnode5.south) {$s_{t+1}$};
    
        % edges
        \draw[->] (supnode1) -- (supnode5) node[midway, above] {$a_t = a_t^{1:M}$};
    
        % rewards
        \draw[->] (supnode1.north east) -- ++(45:0.3) node[right, above] {$r_{t-1}$}; 
        \draw[->] (supnode5.north east) -- ++(45:0.3) node[right, above] {$r_t$};
        
        %%%%%%%%%%%%%%%%%%%%%%%%%%%%%%%%%%%
        % the parallel MDP
        \node[circle,draw, line width=0.8mm] (parnode1) at (0,1.5) {};
        \node[circle,draw, line width=0.8mm] (parnode5) at (12,1.5) {};
    
        % labels
        \node[below] at (parnode1.south) {$s_t$};
        \node[below] at (parnode5.south) {$s_{t+1}$};
    
        % edges
        \draw[->] (parnode1) -- (parnode5) node[midway, above] {$a_t^m$};
    
        % rewards
        \draw[->] (parnode1.north east) -- ++(45:0.3) node[right, above] {$r_{t-1}$}; 
        \draw[->] (parnode5.north east) -- ++(45:0.3) node[right, above] {$r_t$};
    
        % unobserved actions
        \draw[<-] (parnode5.north west) -- ++(135:0.4) node[left, above] {$a^{-m}_t$};
    
        %%%%%%%%%%%%%%%%%%%%%%%%%%%%%%%%%%%
        % the sequential MDP
        \node[circle,draw, line width=0.8mm] (node1) at (0,0) {};
        \node[circle,draw] (node2) at (3,0) {};
        \node[circle,draw] (node3) at (6,0) {};
        \node[circle,draw] (node4) at (9,0) {};
        \node[circle,draw, line width=0.8mm] (node5) at (12,0) {};
        % node labels
        \node[below] at (node1.south) {$\underbrace{[s_t,]}_{s^0_t}$};
        \node[below] at (node2.south) {$\underbrace{[s_t, a_t^1]}_{s^1_t}$};
        \node[below] at (node3.south) {$\underbrace{[s_t, a_t^{1:2}]}_{s^2_t}$};
        \node[below] at (node4.south) {$\underbrace{[s_t, a_t^{1:M-1}]}_{s^{M-1}_t}$};
        \node[below] at (node5.south) {$\underbrace{[s_{t+1},]}_{s^0_{t+1}}$};
        % the MDP edges
        \draw[->] (node1) -- (node2) node[midway, above] {$a_t^1$};
        \draw[->] (node2) -- (node3) node[midway, above] {$a_t^2$};
        \draw[dashed, ->] (node3) -- (node4);
        \draw[->] (node4) -- (node5) node[midway, above] {$a_t^M$};
        % the MDP rewards
        \draw[->] (node1.north east) -- ++(45:0.3) node[right, above] {$r_{t-1}$};
        \draw[->] (node2.north east) -- ++(45:0.3) node[right, above] {$0$};
        \draw[->] (node3.north east) -- ++(45:0.3) node[right, above] {$0$};
        \draw[->] (node4.north east) -- ++(45:0.3) node[right, above] {$0$};
        \draw[->] (node5.north east) -- ++(45:0.3) node[right, above] {$r_t$};
    \end{tikzpicture}
}
\caption{Transitions in the original MDP (upper), reformulated as a parallel MDP (pMDP, middle) and a sequential MDP (sMDP, below, adapted from \cite{metz2019discrete}), where $[\cdot,\cdot]$ represents augmented state vectors.}
\label{fig:smdp}
\end{figure}
In this work we aim at solving Markov Decision Processes (MDP) with discrete action spaces, which can be defined as $\mathcal{M} = (\mathcal{S}, \mathcal{A}, P, R)$ where $\mathcal{S} \subseteq \mathbb{R}^N$ is the state space. $\mathcal{A} = \mathcal{A}_1 \times ... \times \mathcal{A}_M$ is a factorization of $\mathcal{A} \subset \mathbb{N}^M$. 

\noindent\textbf{Parallel MDP.} Given the factorization of $\mA$, we define a parallel MDP (pMDP) $\mM$ as an instance of a stochastic or Markov game $\mG = (\mathcal{S}, \mathcal{A}, P, R)$, where $\mA$ is factorized per definition. Each $\mA_m$ represents the actions available to one of the $M$ players. $R$ defines a pay-off function per agent (e.g. $R: \mS \times \mA \rightarrow \mathbb{R}^M$) \citep{LeytonBrown2008EssentialsOG} but we assume all agents to receive a shared reward to solve the original MDP $\mM$. The distributions over the next state $S_{t+1}$ and reward $R_t$ not only depend on its own action but also the actions selected by all other players, denoted as $a^{-m}_t$. However in the pMDP these actions go unobserved.

\noindent\textbf{Sequential MDP.} Alternatively, given the factorization we define a sequential MDP (sMDP). We introduce intermediate steps into $\mM$ which correspond to selecting one action after the other: $\mathcal{M}_\text{seq} = (\mS_\text{seq}, \mA_\text{seq}, P_\text{seq}, R_\text{seq})$, where $\mS_\text{seq} = \langle \mS_0, \mS_1, ... \mS_{M-1} \rangle$ and $\mathcal{A}_\text{seq} = \langle \mA_1, ... , \mA_M \rangle$  are ordered sets with $\mS_0 = \mS$ and $\mS_m = \mS \times \mA_1 \times ... \times \mA_{m}$. Transitions happen periodically $\mS_m \rightarrow \mS_{(m+1) \text{mod} M}$, with reward given and time $t$ incremented upon transitions into $s^0 \in \mS_0$. In state $s^m \in \mS^m$ action $a^{m+1}$ can only be selected  from $\mA_{m+1}$ \citep{metz2019discrete}.

\subsection{Background on (D)DQN} \label{app:background_dqn}
Before we translate the MDP reformulations into our modifications to DDQN \citep{vanhasselt2015ddqn} we want to provide a brief introduction to Q-Learning and Deep Q-Networks and recommend the excellent text-book by \cite{SuttonBarto18} for a more thorough introduction.
\paragraph{Q-Learning and TD-Updates} Q-Learning \citep{watkins1992qlearning} is a widely used approach to learn optimal policies in MDPs  by learning the state-value function $Q: \mS \times \mA \rightarrow \mathbb{R}, (s,a) \mapsto Q(s,a)$. $Q(s,a)$ is an estimate of the accumulated (and potentially discounted) reward obtainable over an entire episode if choosing action $a$ in state $s$. Given $Q$ we define our policy $\pi(s) = \arg \max_a Q(s,a)$ and analogously we can compute the value function $V: \mS \rightarrow \mathbb{R}, s \mapsto V(s)$ of a state as $V(s) = \max_a Q(s,a)$. The idea of Q-Learning is to update our estimate $Q$ using temporal-difference (TD) updates through the Bellman Optimality Equation: $Q(s_t,a_t) = R_t + \gamma \max_a Q(S_{t+1}, a)$. $R_t$ and $S_{t+1}$ are random variables for the reward and next state we will end up with, if taking action $a_t$ in state $s_t$. Given an observed transition $(s,a,r,s')$ in our MDP the TD-update of $Q$ is:
\begin{align}
    Q(s,a) \leftarrow Q(s,a) + \alpha [r + \gamma \max_{a'} Q(s', a') - Q(s,a)] = Q(s,a) + \alpha [r + \gamma V(s') - Q(s,a)]
\end{align}
Hence we define $r + \gamma \max_{a'} Q(s',a') = r+\gamma V(s')$ as our TD-target. 
\paragraph{Q-Learning Through Function Approximation Using (D)DQN} In discrete state and action spaces of very limited size we might learn tabular entries per pair $(s,a) \in \mS \times \mA$. For very big or infinite state and/or action spaces we have to resort to function approximation, for example through Deep Q-Networks (DQN, \cite{mnih2015dqn}). Here we update the parameters $\theta$ of our Q-function by minimizing the loss:
\begin{align}
    \mathcal{L(\theta)} = \mathbb{E}_{(s,a,r,s') \sim D}[(r + \gamma \max_{a'} Q(s',a';\theta^-) - Q(s,a;\theta))^2]
\end{align}
Parameters $\theta^-$ represent target networks that are updated with a delay. $D$ is a replay buffer where we store transitions (experiences) we collect while interacting with the environment. To update our Q-network we use mini-batches sampled from $D$. Analogous to Q-Learning the TD-target for DQN is $r + \gamma \max_{a'} Q(s',a'; \theta^-)$. We also note that in DQN Q-networks are mappings $q_\theta: \mS \rightarrow \mathbb{R}^{|\mA|}$. That is given a state $s$ they assign a value to each action $a$ or action combination in action vector $\mathbf{a} = a^{1:M} \in \mathbb{N}^M$ in case of $M$-dimensional action spaces. This means $Q(s,a; \theta) = q_\theta(s)[a]$.\\
Instead of the original DQN we implemented our approaches using Double DQN (DDQN, \cite{vanhasselt2015ddqn}) which is a minor extension to DQN but does not affect the presented conceptualization.

\subsection{Learning Q-Networks in the MDP Reformulations} \label{app:dqn_in_mdp}
\begin{table}[h]
    \def\arraystretch{1.5}
    \centering
    \begin{tabular}{|l|l|l|}
    \hline
    \textbf{Algorithm} & \textbf{Underlying MDP} & \textbf{Q-network(s) to learn} \\
    \hline
    DDQN & original MDP & $q: \mS \rightarrow \mathbb{R}^{|\mA|}$\\
    \hline
    IQL & parallel MDP & $\{q^m: \mS \rightarrow \bR^{|\mA_m|} \mid m = 1, ..., M\}$\\
    \hline
    SAQL/simSDQN & sequential MDP & $\{q^m: \mS \times_{i=1}^{m-1} \mA_i \rightarrow \bR^{|\mA_m|} \mid m = 1, ..., M\}$\\
    \hline
    \end{tabular}
    \caption{Q-networks to learn in our evaluated algorithms.}
    \label{tab:q_networks}
\end{table}
In this section, we discuss in detail how to learn policies for different reformulations of the original MDP (Figure \ref{fig:smdp}). For ease of presentation, we drop references to $\theta$. Solving the original MDP is straightforward and requires the learning of a single policy. The related Q-network can be updated against the common TD-target.

For the parallel MDP, we apply Independent Q-Learning (IQL) \citep{tampuu2015iql}, learning $M$ policies, one per action dimension. Each associated Q-network takes the original state $s_t$ as input and maps it to the Q-values of its respective action dimension. Since the actions of other agents go unobserved, they can be treated as unobserved (and nonstationary) environment dynamics. Consequently, we can update each of the $M$ Q-networks independently using the shared, common reward, and the usual TD-target.

In the sequential MDP \citep{metz2019discrete}, policies can observe not only the current state $s_t$, but also the action dimensions already selected for that state. Accordingly, the Q-networks learn to map these augmented observations to the action values of their respective dimensions. We implemented two approaches to update these Q-networks, with different underlying interpretations.
The first approach, Sequential Agent Q-Learning (SAQL), views the sequential MDP as a sequential stochastic game. In substate $s_t^{m-1} = [s_t,a^{1:m-1}]$, it is agent $m$'s turn to choose an action, observing the actions already taken by other players in round $t$. Agent $m$ will make its next observation in round $t+1$, with some fellow players having already acted. Here, we apply the standard DDQN algorithm, limited to the respective action dimension, using an augmented observation space and the shared reward $r_t$, since the game setting is fully cooperative. IQL can be seen as an ablation of SAQL, with identical Q-networks and TD-targets, except for excluding the observation of other agents' actions.

The second approach, Simplified Sequential DQN (simSDQN), is mainly identical to the original proposal by \citet{metz2019discrete}. Our modification is to omit the upper Q-network, because we couldn't successfully train under this setup. We refer the reader to the paper by \cite{metz2019discrete} for the role of the upper Q-network. In simSDQN, we explicitly solve the sequential MDP by updating our Q-Functions against the state-value of the next substate in the sMDP. Using $M$ Q-networks can be viewed as tabular entries for each of the $M$ substates recurring periodically. Table \ref{tab:q_networks} lists the Q-networks to be learned for our different approaches, and Equation \eqref{eq:td_targets} formalizes the targets for the Q-network updates.
\begin{subequations}
\begin{align}
    \text{target}_{\text{DDQN}} & = r_t + \gamma \max_\mathbf{a} Q(s_{t+1}, \mathbf{a}) = r_t + \gamma V(s_{t+1}) \\
    \text{target}_{\text{IQL}}^m & = r_t + \gamma \max_{a^m} Q^m(s_{t+1}, a^m) = r_t + \gamma V^m(s_{t+1}) \\
    \text{target}^m_{\text{SAQL}} & = r_t + \gamma \max_{a^m} Q^m([s_{t+1}, a_{t+1}^{1:m-1}], a^m) = r_t + \gamma V^m(s^m_{t+1}) \\
    \text{target}^m_{\text{simSDQN}} & = \begin{cases}
                                    \max_{a^{m+1}} Q^{m+1}([s_t, a^{1:m}_t], a^{m+1}) = V^{m+1}(s^{m+1}_t), & \quad 1 \leq m \leq M-1 \\
                                    r_t + \gamma \max_{a^1} Q^{1}([s_{t+1},], a^1) = r_t + \gamma V^{1}(s^{0}_{t+1}), & \quad m = M
                              \end{cases}
\end{align} \label{eq:td_targets}
\end{subequations}

\section{Hyperparameter Settings} \label{app:hyperparams}
\begin{table}[h]
\centering
\begin{tabular}{|l|c|c|c|c|}
\hline
\textbf{Hyperparameters} & \textbf{DDQN} & \textbf{IQL} & \textbf{SAQL} & \textbf{simSDQN} \\
\hline
Learning Rate ($\alpha$) & 1.0076e-4 & 3.2680e-5 & 7.7590e-5 & 3.0855e-4 \\
\hline
Discount Factor ($\gamma$) & 0.9349 & 0.9147 & 0.9086 & 0.9696 \\
\hline
Start Exploration Rate ($\epsilon_\text{start}$) & 0.2382 & 0.9289 & 0.4607 & 0.1341 \\
\hline
Target Update Frequency & 15 & 39 & 12 & 33 \\
\hline
Target Soft Update Factor ($\tau$) & 0.2613 & 0.1258 & 0.6196 & 0.4765 \\
\hline
Batch Size & 220 & 63 & 67 & 105 \\
\hline
End Exploration Rate ($\epsilon_\text{end}$) & \multicolumn{4}{c|}{0.01} \\
\hline
Exploration fraction (linear $\epsilon$ - decay) & \multicolumn{4}{c|}{0.5} \\
\hline
Optimizer & \multicolumn{4}{c|}{Adam\footnote{untuned apart from $\alpha$, \texttt{torch 2.1.0} defaults}} \\
\hline
Replay Buffer Size & \multicolumn{4}{c|}{2500} \\
\hline
\end{tabular}
\caption{Hyperparameters per method, used for our experiments.}
\label{table:1}
\end{table}

\section{Policy architecture} \label{app:policy_architecture}
We represented the Q-functions of our learned policies as $3$ layer MLPs with ReLU activation functions. Note that for the factorized policies IQL, SAQL, simSDQN the number of policies to learn corresponds to the dimension of the benchmark.
For all policies, we used shared numbers of hidden units: $120$ and $84$ in the first and second hidden layers, respectively. For DDQN and IQL the number of input units corresponds to the dimensionality of the observation space of the benchmark environment. For sequential policies SAQL and simSDQN the size of the input layer of $Q^m$ is $\texttt{dim\_observation\_space} + m$ for $m \in \{0, ..., M-1\}$ (see Appendix \ref{app:pl}). The output size for all factorized policies is the number of actions per action dimension $\texttt{n\_act} \in \{3,5,10\}$, for DDQN it is the number of all possible action combinations over all action dimensions $\texttt{n\_act}^M = \texttt{n\_act}^\texttt{dim}$.

\section{Compute Resources} \label{app:resources}
Unless otherwise stated, the experiments were run on nodes using a single CPU "Intel Xeon Gold 6230".
For DDQN experiments with higher-dimensional action spaces ($\texttt{dim}=10$) or more discrete action choices per action dimension ($\texttt{n\_act} = 10$), the resulting Q-networks were substantially larger, necessitating the use of GPU accelerators. These experiments were executed on nodes equipped with "NVIDIA Tesla V100" GPUs.

\section{Different Importance Decays} \label{app:diff_importance}

\begin{figure} [h]
    \centering
    \includegraphics[width=\textwidth]{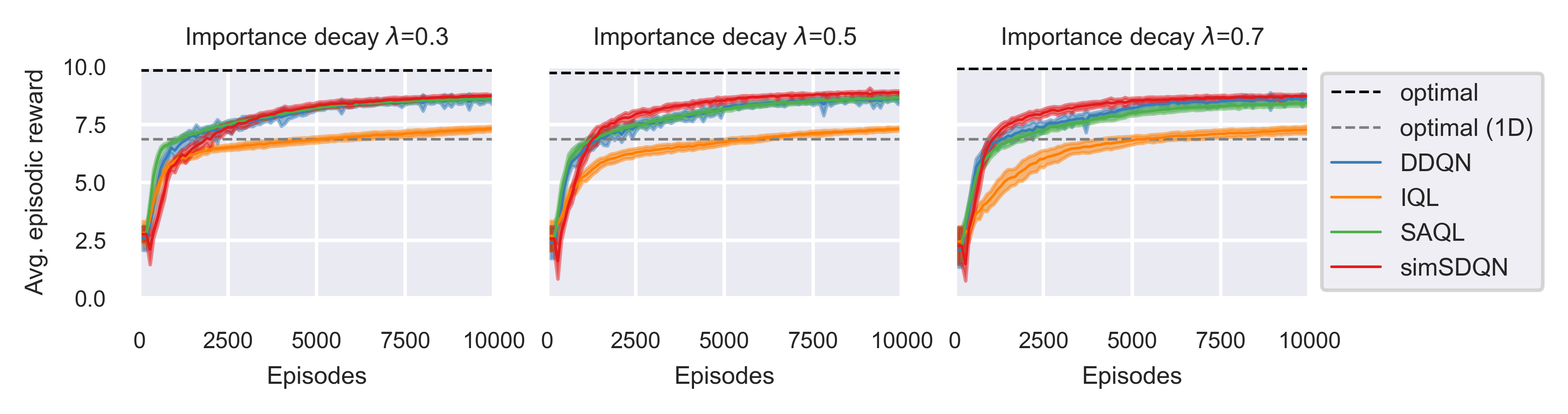}
    \caption{Average episodic test rewards (mean, std from 20 seeds) on the 5D Piecewise Linear benchmark, for different importance decays $\lambda$. Lower $\lambda$ means importance is decreased more strongly from action dimension to action dimension.}
    \label{fig:diff_importance}
\end{figure}

\section{Reversed Importances} \label{app:reverse_importances}
\begin{figure}[h]
    \centering
    \includegraphics{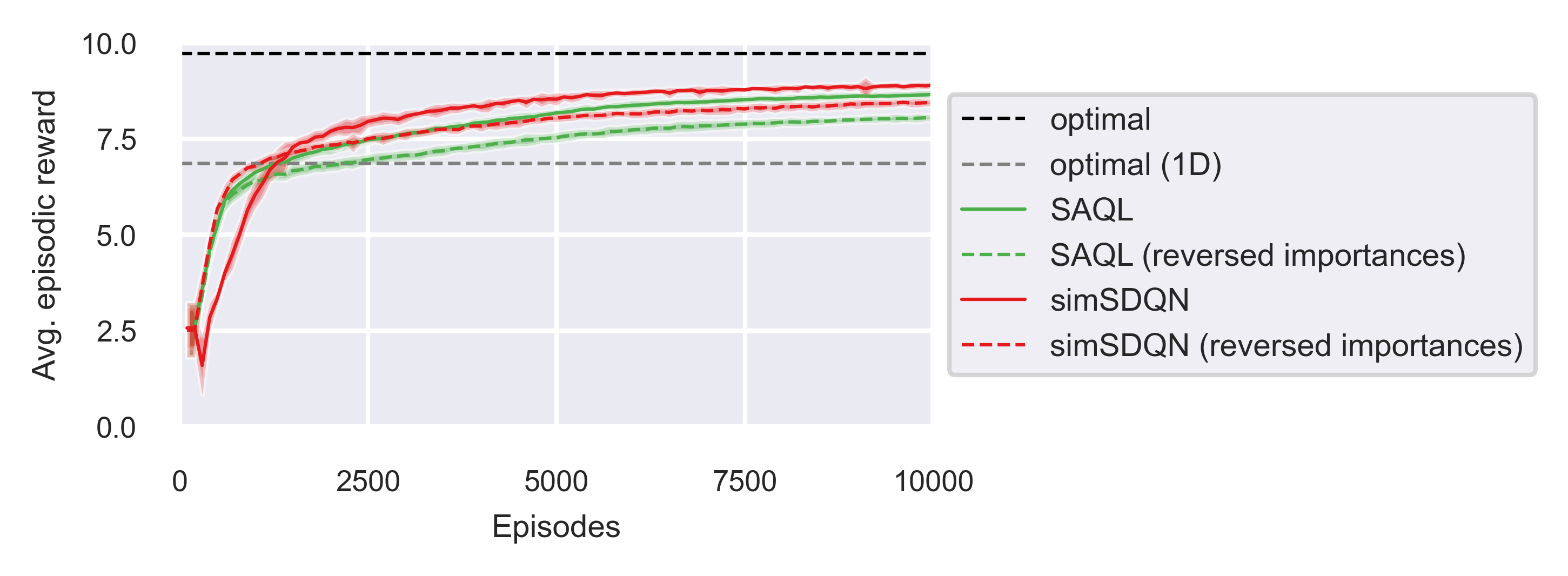}
    \caption{Average episodic test rewards (mean, std from 20 seeds) obtained by SAQL and simSDQN when selecting actions with descending or ascending (reversed) importance on 5D Piecewise Linear benchmark.}
    \label{fig:reverse_imp}
\end{figure}
The ablation presented in Figure \ref{fig:reverse_imp} confirmes our intuition regarding selecting actions in descending order of their importance for sequential policies and emphasizes the signficance of getting the order of importances right. This analysis also sheds light on a potential issue of simSDQN when facing higher dimensonal action spaces: In our design, rewards are only assigned when performing the TD-update for the Q-network responsible for the least important action dimension. This requires the propagation of reward information to more important action dimensions (for a conceptual illustration, refer to Figure \ref{fig:smdp}). By reversing the order of action dimensions, the Q-network corresponding to the most important action dimension can be directly updated towards the reward signal. This leads to a noticeable speed up of learning in the initial training phase.

% ==== Bibliography
% print bibliography -- for bibtex / natbib, use:

% \bibliography{...}

% and for biber / biblatex, use:

% \printbibliography

% supplemental material -- everything hereafter will be suppressed during
% submission time if the hidesupplement option is provided!

\end{document}